\documentclass[letterpaper, 10pt, journal, twoside]{IEEEtran}
\usepackage{amsmath,amsfonts}
\usepackage{algorithmic}
\usepackage{algorithm}
\usepackage{array}
\usepackage[caption=false,font=normalsize,labelfont=sf,textfont=sf]{subfig}
\usepackage{textcomp}
\usepackage{stfloats}
\usepackage{url}
\usepackage{verbatim}
\usepackage{graphicx}
\usepackage{cite}

\usepackage{tikz}
\usetikzlibrary{positioning, fit, arrows.meta, calc} 
\usepackage{svg}
\usepackage[super]{nth}
\usepackage{color}
\usepackage{here}

\begin{document}

\title{Observation-Augmented Contextual Multi-Armed Bandits for Robotic Search and Exploration}

\author{Shohei Wakayama and Nisar Ahmed$^*$
\thanks{Manuscript received: February 21, 2024; Revised: July 21, 2024; Accepted: August 15, 2024.}
\thanks{This paper was recommended for publication by Editor Giuseppe Loianno upon evaluation of the Associate Editor and Reviewers' comments.}
\thanks{$^*$The authors are with the Smead Aerospace Engineering Sciences Department, University of Colorado Boulder, Boulder, CO 80303 USA {\tt\small [shohei.wakayama;nisar.ahmed]@colorado.edu}}
\thanks{Digital Object Identifier (DOI): see top of this page.}
}

\markboth{IEEE Robotics and Automation Letters. Preprint Version. Accepted August, 2024}%
{Wakayama and Ahmed: Observation-Augmented Contextual Multi-Armed Bandits}


\maketitle

\newtheorem{definition}{Definition}
\newtheorem{problem}{Problem}
\newtheorem{example}{Example}
\newtheorem{remark}{Remark}
\newtheorem{lemma}{Lemma}

\newcommand{\pa}[1]{\textcolor{magenta}{[PA: #1]}}
\newcommand{\sw}[1]{\textcolor{red}{[SW: #1]}}
\newcommand{\ml}[1]{\textcolor{blue}{[ML: #1]}}
\newcommand{\na}[1]{\textcolor{magenta}{[NA: #1]}}
\newcommand{\qh}[1]{\textcolor{magenta}{[QH: #1]}}
\newcommand{\zl}[1]{\textcolor{cyan}{[ZL: #1]}}
\newcommand{\edt}[1]{\textcolor{blue}{#1}}

\newcommand{\reals}{\mathbb{R}}
\newcommand{\naturals}{\mathbb{N}}
\newcommand{\N}{\mathcal{N}}
\newcommand{\C}{\mathcal{C}}
\renewcommand{\L}{\mathcal{L}}
\newcommand{\needcite}{\textbf{CITE} } 
\newcommand{\dist}{\mathcal{D}}
\newcommand{\scurr}{s_K}
\newcommand{\distest}{\tilde{\mathcal{D}}}
\newcommand{\expec}[1]{\mathbb{E}[#1]}
\newcommand{\variance}[1]{\text{var}[#1]}
\newcommand{\entropy}[1]{\mathbf{H}[#1]}
\newcommand{\st}{\text{ s.t. }}
\newcommand{\cvec}{\mathbf{c}}
\newcommand{\ccvec}{\mathbf{C}}
\newcommand{\cvecest}{\tilde{\mathbf{c}}}
\newcommand{\ccvecest}{\tilde{\mathbf{C}}}
\newcommand{\kld}{D_{KL}}
\newcommand{\planparams}{\theta_{s_c, \pi^*}}
\newcommand{\planset}{A^*}
\newcommand{\actiontxt}[1]{\texttt{#1}}

\newcommand{\mean}{\mathbf{\mu}}
\newcommand{\cov}{\mathbf{\Sigma}}

\newcommand{\DTS}{\text{DTS-sc}\xspace}
\newcommand{\pref}{\mathrm{pr}}

\newcommand{\Q}{{Q_\phi}}

\newcommand{\Rnine}{\textcolor{teal}{\textbf{R9}}}
\newcommand{\Rsix}{\textcolor{violet}{\textbf{R6}}}

\begin{abstract}
We introduce a new variant of contextual multi-armed bandits (CMABs) called observation-augmented CMABs (OA-CMABs) wherein a robot uses extra outcome observations from an external information source, e.g. humans. 
In OA-CMABs, external observations are a function of context features and thus provide evidence on top of observed option outcomes to infer hidden parameters. However, if external data is error-prone, measures must be taken to preserve the correctness of 
inference. 
To this end, 
we derive a robust Bayesian inference process for OA-CMABs based on recently developed probabilistic semantic data association techniques, 
which 
handle complex mixture model parameter priors and hybrid discrete-continuous observation likelihoods for semantic external data sources. 
To cope with combined uncertainties in OA-CMABs, we also derive a new active inference algorithm for optimal option selection based on approximate expected free energy minimization. This generalizes prior work on CMAB active inference 
by accounting for faulty observations and non-Gaussian distributions. 
Results for a simulated 
deep space search site selection problem show that, even if incorrect semantic observations are provided externally, e.g. by scientists, efficient decision-making and robust parameter inference are still achieved in a wide variety of 
conditions.
\end{abstract}


\section{Introduction} \label{sec: introduction}
\IEEEPARstart{I}{n} uncertain remote environments, such as deep space and underwater, there are many situations in which an autonomous robot must choose the best option among multiple alternatives to accomplish a task. 
Such decisions are challenging 
since
the world is stochastic, and the distributions of outcomes resulting from options often are unknown {\it a priori}. One efficient way to reason about the parameters of outcome distributions would be to utilize extra observations from external information sources, e.g. humans and object detectors \cite{redmon2016you}. Yet, if these extra 
observations are noisy and a robot naively uses them to infer parameters, then such 
observations can do more harm than good unless these additional uncertainties can be mitigated. 
\begin{figure}[t]
    \centering
    \centerline{\includegraphics[width=\columnwidth]{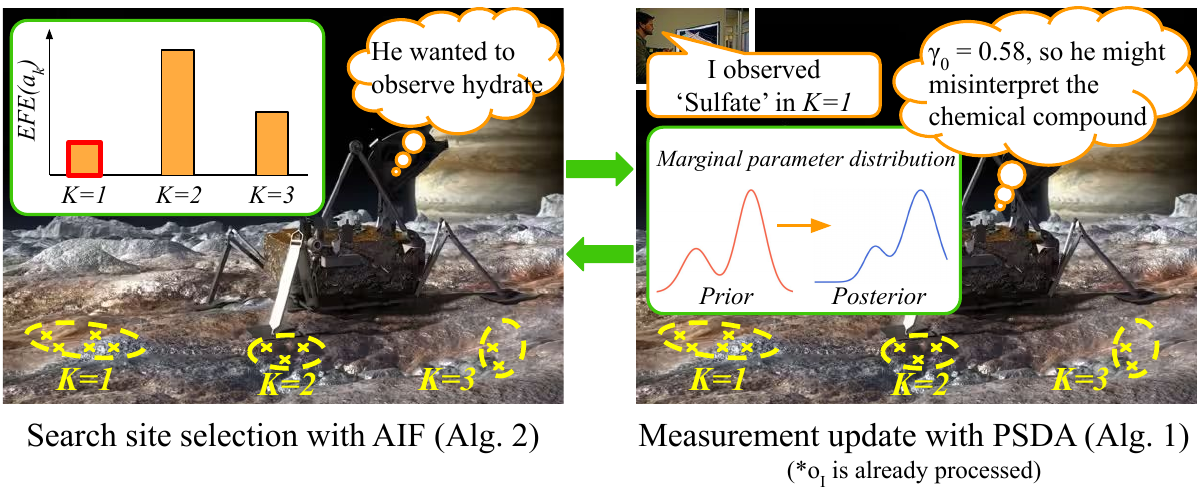}}
    \caption{Asynchronous collaborative coupled decision-making and state estimation scenario: a deep space robotic lander iteratively selects the best search site (left) and updates the estimates with its own sensor measurements and delayed (possibly erroneous) 
    scientist observations (right).}
    \label{fig: scenario}
    \vspace{-0.3in}
\end{figure}

As an example, consider the motivating deep space exploration scenario illustrated in Fig. \ref{fig: scenario}. A robotic lander dispatched to an icy 
moon must identify the best scientific exploration site with geological features of interest to scientists back on Earth using sensors with relatively low energy consumption, such as laser-induced breakdown spectroscopy (LIBS) \cite{jolivet2019review}. If the mission time were unlimited, the robot would be able to make a sufficient number of observations at all candidate digging sites to accurately estimate their geological characteristics and identify the best site for risky operations like drilling, digging, etc. (e.g. the number of measured points $\times$ in each search site is considerable). However, in reality, 
external factors (e.g. high radiation) 
make it necessary to perform the identification as quickly and efficiently as possible \cite{adam2016europa}. Hence, a balance must be struck between increasing the certainty of plausible search sites (i.e. {\it exploitation}) and reducing the uncertainty of less well-studied search sites (i.e. {\it exploration}). 
The robot can make use of 
contextual information from the surrounding environment, for instance, the albedo of search sites. Furthermore, in addition to (noisy) LIBS readings for determining specimen categories at each site, the robot may be able to receive supplementary discrete semantic observations from Earth-bound human scientists on a time delay (e.g. ``Site \#2 contains a lot of sulfate,
site \#4 has very few salts"). 

Previous work formulated such robotic exploration scenarios as contextual multi-armed bandits (CMABs) \cite{steve2019,wakayamaICRA2023}. In CMABs, for each option, there exist hidden parameters that inform the robot of  expected outcomes for taking specific options as a function of side information, i.e. {\it contexts}. Typically, prior distributions for estimating these parameters are updated via Bayes' rule based on the outcomes {\it only} obtained by executing options. 
The outcome observations are usually derived from onboard sensor data and are assumed {\it a priori} to have known relevance to the parameters, 
which is modeled by a probabilistic likelihood function. However, it is not known in advance how likely it is that any extra outcome observations provided by an external source (e.g. scientist in Fig. \ref{fig: scenario}) will be useful for hidden parameter estimation. This is because humans can make mistakes, and so extra care must be taken 
to ensure data validity for inference and decision making. 
This makes both processes more complex for CMABs. 

Hence, in this study, we first introduce a new variant of CMABs called {\it observation-augmented CMABs (OA-CMABs)} wherein a decision-making robot can utilize extra outcome observations from an external information source. Then, we enable 
a robust measurement update step by adapting
probabilistic semantic data association (PSDA), which was developed for 
human-robot collaborative sensing \cite{wakayamaTRO} to deal with data validation problems like those described above. This adaptation is a novel approach to CMAB-based decision making. 
Additionally, we introduce a generalized way to derive expected free energy (EFE) for active inference \cite{smith2022step}, which was recently shown to be a powerful approach for 
determining option selection strategies in bandit problems \cite{markovic2021,wakayamaICRA2023}.  Our generalization accounts for hidden parameters modeled via mixture distributions, which naturally appear when 
using data validation techniques like PSDA. These methods are validated in a simulated collaborative asynchronous search site selection scenario. The results show that using the PSDA measurement update process with generalized EFE-based active inference identifies the best site  with fewer iterations than conventional baselines, leading to smaller cumulative regrets even when incorrect human semantic observations are provided to robots. 

For the remainder of the paper: Sec. \ref{sec: background} provides an overview of CMABs, data association, and active inference. Sec. \ref{sec: methodologies} then describes the problem setup and PSDA measurement update process, and explains how to compute the EFE for mixture prior distributions. In Sec. \ref{sec: simulation_study}, the simulation setup and the results are presented, followed by conclusions and future perspectives in Sec. \ref{sec: conclusions}.

\section{Background} \label{sec: background}
\subsection{Contextual Multi-Armed Bandits} \label{sec: cmab} 
A CMAB is a reinforcement learning problem which iterates between two steps: inference over latent parameters that enable prediction of expected outcomes (based on prior outcomes), and option selection given available parameter estimates.
Although apparently simple compared to other sequential decision making problems, 
such as MDPs \cite{kochenderfer2015decision} and POMDPs \cite{kurniawati2022partially}, the scope of CMAB applications is broad and long studied in areas such as recommendation systems and finance \cite{bouneffouf2020}. 

CMABs 
typically aim to minimize cumulative regret, i.e. the difference between optimal and selected option outcomes. 
Bayes' theorem is primarily used to infer latent predictive model parameters between option selections. However, this requires adequate models of uncertainties, including those arising from 
observations obtained by the agent. Whereas such observations are usually available in bandit settings only after options are selected, the problem of handling uncertainties from observations available outside of option selection is examined here for the first time and addressed in Sec. \ref{sec: data_association}. For option selection, $\varepsilon$-greedy, strategies based on the upper confidence bound (UCB) \cite{auer2002}, and Thompson sampling (TS) \cite{thompson1933} are well-known, but often require many iterations or heuristics to achieve good performance. More recently, active inference-based option selection methods have been shown to identify the best option with fewer iterations, though in certain cases 
they may potentially get stuck in local minima \cite{markovic2021,wakayamaICRA2023}.

\subsection{Active Inference and Expected Free Energy} \label{sec: aif_efe}
Active inference is a neuro-inspired decision-making framework that applies the free energy principle \cite{friston2010} 
to the behavioral norms of biological agents. 
In active inference theory \cite{smith2022step}, agents are thought to select an action/option that minimizes a quantity called expected free energy (EFE), which has attracted interest in computational neuroscience and robotics \cite{lanillos2021} as a mechanism for autonomous sequential decision making under uncertainty. Agents reasoning via the EFE consider not only the utility gained by executing an option, but also how much uncertainty about a hidden state can be reduced by executing that option, thus naturally balancing exploitation and exploration. This is achieved by reasoning against an {\it evolutionary prior} (a.k.a. prior preference), which defines an outcome distribution that the agent expects to see when undertaking options and which provides a reference for updating the agent's internal (probabilistic) world model. 

Recent work in \cite{wakayamaICRA2023} developed 
techniques for calculating EFE in CMABs with semantic observations using variational and Laplace approximation for Bayesian parameter updating. 
However, like other work in the bandit study \cite{markovic2021}, active inference does not account for problems where agents must also reason about the validity of the observations that they rely on to perform 
parameter inference. This not only requires more advanced
inference mechanisms and probability distribution representations to account for multiple data validity hypotheses, but also complicates the calculation of EFE for active inference (as this can be a multimodal/non-convex function).

\subsection{Reasoning about Data Validity } \label{sec: data_association}
When utilizing observations from external information sources to accelerate the estimates of unknown parameters, care should be taken to assess their validity. 
This is because if incorrect observations are naively used to derive parameter posteriors, the resulting posteriors can diverge significantly from true parameter values and lead to suboptimal decision making behavior. Although there are several possible ways to validate the external observations \cite{breck2019data,horvat2020impact}, in this study we focus on {\it data association (DA)} \cite{bar-shalom2009}, a class of estimation methods that explicitly accounts for observation origin uncertainties. Prominent DA methods include nearest neighbor \cite{li1996tracking}, probabilistic data association (PDA) \cite{bar-shalom2009}, and multi-hypothesis tracking \cite{kim2015multiple}, all of which differ mainly in how they handle data association variables, which explain origin hypotheses. 

Among these, we employ Probabilistic Semantic Data Association (PSDA) \cite{wakayamaTRO}, which fits the scheme of CMABs that update the distribution of hidden parameters based on semantic observations obtained at {\it every} decision-making iteration. Note that semantic observations refer to noisy categorical descriptions of abstract object/event properties or relations. 
Such data has attracted attention in robotics due to the usefulness of meaningfully grounded reasoning over continuous variables such as spatial positions \cite{tse2018}.

\section{Methodology} \label{sec: methodologies}
\subsection{Problem Statement} \label{sec: problem_statement}
We first formulate the robotic decision-making and inference problem under uncertainty introduced in Sec. \ref{sec: introduction} as a contextual bandit (CMAB). Suppose the total number of options (i.e. search sites) taken into account 
is $K$. Note that these options are equivalent to the bandit arms and selecting an option $k\in\{1,\cdots,K\}$ is denoted as $a\!=\!k$ (for the shorthand notation, in the following, we use $a_k \leftrightarrow a\!=\!k$). Also, suppose that 
a semantic observation, $o_k$ of each option $k$ from an observation source (i.e. the robot sensor or human input)
is multicategorical across $F$ labels, i.e. $o_k\!=\!f, f \in 
{\cal F} = \{1, \cdots, F\}$. Thus, the probability that a feature $f$ is observed by investigating option $k$ can be described as the softmax likelihood function \cite{bishop2006,nisar2018}, 
{\allowdisplaybreaks
\abovedisplayskip = 2pt
\abovedisplayshortskip = 2pt
\belowdisplayskip = 1pt
\belowdisplayshortskip = 1pt
\begin{align} \label{eq: softmax}
    p(o_k=f|\vec{\Theta}_k;\vec{x}_c, \vec{x}_k) = \frac{e^{\vec{\Theta}_{k,f}^T (\vec{x}_c + \vec{x}_k)}}{\sum_{h=1}^{F} e^{\vec{\Theta}^T_{k,h}(\vec{x}_c + \vec{x}_k)}},
\end{align}
}where $\vec{\Theta}_k\!=\![\vec{\Theta}_{k,1}, \cdots, \vec{\Theta}_{k,F}], \vec{\Theta}_k \in \mathbb{R}^{C\times F}$ is a hidden linear parameter vector unique to the option $k$, and $\vec{x}_c, \vec{x}_k \in \mathbb{R}^{C}$ 
are the option-agnostic and the option-specific context vectors\footnote{Without loss of generality, the context vectors are assumed common for all iterations and a bias term can be ignored. 
}, where $C$ is the context feature dimension. 

Recall, the CMAB objective is to minimize cumulative regrets. 
Here, a unit reward ($1$) is provided if a 
preferable feature $f_p \in {\cal F}$ is observed, and no reward ($0$) is given if any other feature is observed. For the search site selection scenario, 
$f_p$ represents a 
chemical
label that scientists want the robot to find. Thus, if the probability of observing $f_p$ with the best (unknown {\it a priori}) option is $\psi^*$, the cumulative regret is 
\cite{auer2002}, 
{\allowdisplaybreaks
\abovedisplayskip = 2pt
\abovedisplayshortskip = 2pt
\belowdisplayskip = 2pt
\belowdisplayshortskip = 0pt
\begin{align}
    \mbox{Regret}(T) = T\psi^* - \sum_{k=1}^K N_T(k) \psi_k,
\end{align}
}where $T$ is the total number of iterations, $N_T (k)$ represents the number of times an option $k$ is executed within $T$ iterations, and $\psi_k$ is the probability that $f_p$ is observed by executing the option $k$. So, to minimize the cumulative regrets, the robot is required to efficiently estimate the set of softmax parameters $\vec{\Theta}_k$ for all $k$ to identify the best option. However, in standard CMABs, {\it only a single} observation is obtained per selected options, therefore a number of iterations are typically necessary before these parameters can be correctly estimated. Yet, as is often the case with robotics, external information sources can also be utilized although they may not be immediately available. Hence, in this study, we consider observation-augmented CMABs (OA-CMABs) such that extra outcome observations on the robot's choice of option could be used. Nevertheless, this could be counterproductive if the data validity of 
is not considered as described in Sec. \ref{sec: data_association}. In the following, we first review the conventional measurement update process for CMABs and its problem when external observations are naively fused, and then introduce a robust measurement update process by accounting for the data validity via PSDA. 
We then consider how to undertake option selection while accounting for data validity uncertainties, using the framework of active inference. \vspace{-0.1in}

\subsection{Semantic Data Association Update} \label{sec: update_sda} 
In the conventional measurement update process, a latent parameter vector $\vec{\Theta}_k$ associated with an option $k$ is, by and large, updated via Bayes' theorem based on the observation $o_k\!=\!f$ obtained by executing that option.
{\allowdisplaybreaks
\abovedisplayskip = 2pt
\abovedisplayshortskip = 2pt
\belowdisplayskip = 2pt
\belowdisplayshortskip = 0pt
\begin{align} \label{eq: posterior}
    p(\vec{\Theta}_k|o_k\!=\!f;\!\vec{x}_c,\! \vec{x}_k)\!=\!\frac{p(o_k\!=\!f|\vec{\Theta}_k;\!\vec{x}_c,\!\vec{x}_k)p(\vec{\Theta}_k)}{\int_{\vec{\Theta}_k} p(o_k\!=\!f|\vec{\Theta}_k;\!\vec{x}_c,\! \vec{x}_k)p(\vec{\Theta}_k) d\vec{\Theta}_k}.
\end{align}
}However, as mentioned previously, when an (external) observation $o_k$ is faulty/erroneous, taking it at face value and deriving the posterior of hidden parameters $\vec{\Theta}_k$ for that option reduces the accuracy of predicted outcome probabilities via (\ref{eq: softmax}). 
Although there are multiple ways to evaluate data validity in such cases, particularly when the faulty measurement probability (FP) rate of the external information sources is known
, the probability that a received external observation is valid (referred to here as a \emph{data association 
hypothesis}) can also be explicitly calculated through a Bayesian update to enable more robust parameter inference. This is achieved here using {\it probabilistic semantic data association (PSDA)} \cite{wakayamaTRO}, 
a theoretical generalization of the aforementioned PDA algorithm widely used for {\it continuous-valued} data validation in 
dynamic target tracking. PSDA is able to dynamically assess the association hypothesis probabilities for {\it semantic} observations at every OA-CMAB decision-making iteration.

\subsubsection{Robust PSDA measurement update}
In the PDA/PSDA methodologies, the probability density function (pdf) of a hidden variable becomes a weighted sum of individual pdfs that reflect the histories of past data associations. Thus, a prior pdf of a hidden linear parameter vector $\vec{\Theta}_k$ for OA-CMABs is expressed by the following mixture distribution 
{\allowdisplaybreaks
\abovedisplayskip = 6pt
\abovedisplayshortskip = 6pt
\belowdisplayskip = 6pt
\belowdisplayshortskip = 6pt
\begin{align}
    p(\vec{\Theta}) = \sum_{u=1}^{M} p(\vec{\Theta}|u) p(u), 
\end{align}
}where $p(\vec{\Theta}|u)$ is {\it a mixand} indexed by $u$, $p(u)$ represent a mixture weight, i.e. how plausible a mixand is to describe the prior, and $M$ is the total number of mixands.
Suppose a latent data association (DA) variable is represented by $\zeta$. Here, since the robot is only required to reason if an external outcome observation is correct or not, $\zeta$ is binary, i.e. $\zeta\!=\!0$ and $\zeta\!=\!1$ indicates that the observation $o$ is incorrect and correct, respectively. Hence, the probabilistic graphical model (PGM) for OA-CMABs when processing external observations with DA is illustrated as in Fig. \ref{fig: pgm}. Note that the option index $k$ and the mixture index $u$ are abbreviated for simplicity. 
\begin{figure}
\centering 
\begin{tikzpicture}
    \node[draw, fill=gray!40, circle, minimum size=1.0cm, font=\fontsize{6.5}{10}\selectfont] (theta^t) {$\vec{\Theta}^t$};
    \node[draw, fill=yellow!40, circle, minimum size=1.0cm, above left=0.2cm and 0.6cm of theta^t, font=\fontsize{6.5}{10}\selectfont] (o_ex^t) {$o_{E}^t$};
    \node[draw, fill=yellow!40, circle, minimum size=1.0cm, above right=0.2cm and 0.6cm of theta^t, font=\fontsize{6.5}{10}\selectfont] (o_in^t) {$o_{I}^t$};
    \node[draw, fill=yellow!40, circle, minimum size=1.0cm, below right=0.1cm and 0.6cm of theta^t, font=\fontsize{6.5}{10}\selectfont] (a_in^t) {$a_{I}^t$};
    \node[draw, fill=yellow!40, circle, minimum size=1.0cm, above=0.4cm of theta^t, font=\fontsize{6.5}{10}\selectfont] (x^t) {$\vec{x}^t$};
    \node[draw, fill=gray!40, circle, minimum size=1.0cm, below left=0.1cm and 0.6cm of theta^t, font=\fontsize{6.5}{10}\selectfont] (zeta^t) {$\zeta^t$};

    \node[draw, fill=gray!40, circle, minimum size=1.0cm, right=2.8cm of theta^t, font=\fontsize{6.5}{10}\selectfont] (theta^t1) {$\vec{\Theta}^{t+1}$};
    \node[draw, fill=yellow!40, circle, minimum size=1.0cm, above left=0.2cm and 0.6cm of theta^t1, font=\fontsize{6.5}{10}\selectfont] (o_ex^t1) {$o_{E}^{t+1}$};
    \node[draw, fill=yellow!40, circle, minimum size=1.0cm, above right=0.2cm and 0.6cm of theta^t1, font=\fontsize{6.5}{10}\selectfont] (o_in^t1) {$o_{I}^{t+1}$};
    \node[draw, fill=yellow!40, circle, minimum size=1.0cm, below right=0.1cm and 0.6cm of theta^t1, font=\fontsize{6.5}{10}\selectfont] (a_in^t1) {$a_{I}^{t+1}$};
    \node[draw, fill=yellow!40, circle, minimum size=1.0cm, above=0.4cm of theta^t1, font=\fontsize{6.5}{10}\selectfont] (x^t1) {$\vec{x}^{t+1}$};
    \node[draw, fill=gray!40, circle, minimum size=1.0cm, below left=0.1cm and 0.6cm of theta^t1, font=\fontsize{6.5}{10}\selectfont] (zeta^t1) {$\zeta^{t+1}$};

    \draw[-Stealth]
        (theta^t)   edge (o_ex^t)
        (theta^t)   edge (o_in^t)
        (a_in^t)    edge (o_in^t)
        (zeta^t)    edge (o_ex^t)
        (x^t)       edge (o_ex^t)
        (x^t)       edge (o_in^t)
        ($(theta^t.west) + (-1.25cm,0)$) edge ($(theta^t.west) + (-1.25cm,0) + 1.2*({cos(0)},{sin(0)})$)
        (theta^t)   edge (theta^t1)
        (theta^t1)  edge (o_ex^t1)
        (theta^t1)  edge (o_in^t1)
        (a_in^t1)   edge (o_in^t1)
        (zeta^t1)   edge (o_ex^t1)
        (x^t1)      edge (o_ex^t1)
        (x^t1)      edge (o_in^t1)
        (theta^t1) to[out=0, in=180] ++(2.0, 0);
    \draw[-Stealth, dashed]
        (theta^t)   edge (a_in^t)
        (x^t)       edge (a_in^t)
        (theta^t1)  edge (a_in^t1)
        (x^t1)      edge (a_in^t1);

    \node[above right=0.5cm and -0.2cm of x^t, font=\fontsize{6.5}{10}\selectfont] {$\zeta$: DA variable, $o$: outcome, $a$: option,};
    \node[above right=0.12cm and 0.0cm of x^t, font=\fontsize{6.5}{10}\selectfont] {$\vec{\Theta}$: latent parameters, $\vec{x}$: contexts};
    \fill [draw=black, fill=none] (0.1,2.7) rectangle (3.95,2.);
\end{tikzpicture}
\caption{{\small{A PGM for OA-CMABs with DA; observable/latent variables are highlighted in yellow/gray. $(\cdot)_{E}$ and $(\cdot)_{I}$ represent {\it external} and {\it internal} observations/actions. 
Context vectors $\vec{x}_c$ and $\vec{x}_k$ are summarized as $\vec{x}$. Dotted lines 
indicate
causality of option selection.}}}
\label{fig: pgm}
\vspace{-0.2in}
\end{figure}
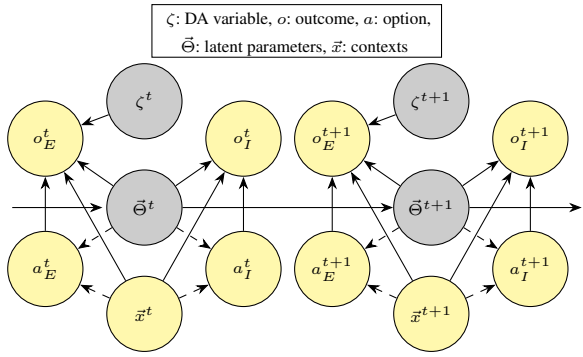In the following, assume that the robot internal sensor data $o_{I}$ is preprocessed, and the external outcome observation $o_{E}$ is denoted as $o$. Given this PGM, the joint posterior of the latent variables $\vec{\Theta}$ and $\zeta$ are described as 
{\allowdisplaybreaks
\abovedisplayskip = 6pt
\abovedisplayshortskip = 6pt
\belowdisplayskip = 6pt
\belowdisplayshortskip = 6pt
\begin{align}
    p(\vec{\Theta}, \zeta|o) = \sum_{u=1}^M p(\vec{\Theta}, \zeta, u|o).
\end{align}
}Although the type of semantic data association posterior can vary depending on how the DA variable $\zeta$ is treated, when the PSDA method is applied, the posterior $p_{PSDA}(\vec{\Theta}|o)$ is 
{\allowdisplaybreaks
\abovedisplayskip = 6pt
\abovedisplayshortskip = 6pt
\belowdisplayskip = 6pt
\belowdisplayshortskip = 6pt
\begin{align} 
    &p_{PSDA}(\vec{\Theta}|o) \nonumber \\
    &\!=\!\sum_{u=1}^M p(\vec{\Theta}, u) \frac{p(\zeta\!=\!0, o)}{\sum_{\zeta} p(\zeta, o)}\!+\!\sum_{u'=1}^M p(\vec{\Theta}, u'|o) \frac{p(\zeta\!=\!1, o)}{\sum_{\zeta} p(\zeta, o)} \label{eq: psda_binary} \\
    &\!=\!\gamma_0 \cdot \sum_{u=1}^M p(\vec{\Theta}, u) + \gamma_1 \cdot \sum_{u'=1}^M p(\vec{\Theta}, u'|o) \label{eq: psda_binary_gamma}
\end{align}
}where $u'$ represents an index of a mixand for the posterior and $\gamma_0\!=\!\frac{p(\zeta=0, o)}{\sum_{\zeta} p(\zeta, o)}$ and $\gamma_1\!=\!\frac{p(\zeta=1, o)}{\sum_\zeta p(\zeta, o)}$ are corresponding to the posteriors of association probabilities. Although the detailed derivation can be referred to \cite{wakayamaTRO}, $p(\zeta\!=\!0,o)$ and $p(\zeta\!=\!1,o)$ in (\ref{eq: psda_binary}) are further expanded as follows.  
{\allowdisplaybreaks
\abovedisplayskip = 2pt
\abovedisplayshortskip = 2pt
\belowdisplayskip = 2pt
\belowdisplayshortskip = 0pt
\begin{align}
    p(\zeta\!=\!0, o) = (1/F) \cdot p(\zeta\!=\!0),
\end{align}
}
{\allowdisplaybreaks
\abovedisplayskip = 2pt
\abovedisplayshortskip = 2pt
\belowdisplayskip = 2pt
\belowdisplayshortskip = 0pt
\begin{align}
    p(\zeta\!=\!1, o) = p(\zeta\!=\!1) \sum_{u=1}^M p(u) \int_{\vec{\Theta}}  p(o|\vec{\Theta}) p(\vec{\Theta}|u) d\vec{\Theta}.
\end{align}
}
The joint posterior of $\vec{\Theta}$ and $u'$ given $o$ in (\ref{eq: psda_binary_gamma}) is 
{\allowdisplaybreaks
\abovedisplayskip = 2pt
\abovedisplayshortskip = 2pt
\belowdisplayskip = 2pt
\belowdisplayshortskip = 0pt
\begin{align}
    &p(\vec{\Theta}, u'|o) = p(\vec{\Theta}|u', o)p(u'|o) \nonumber \\
    &=\!\frac{p(o|\vec{\Theta})p(\vec{\Theta}|u')}{\int_{\vec{\Theta}} p(o|\vec{\Theta})p(\vec{\Theta}|u') d\vec{\Theta}} \cdot \frac{p(u')\int_{\vec{\Theta}}p(o|\vec{\Theta})p(\vec{\Theta}|u')}{\sum_{u'} p(u') \int_{\vec{\Theta}} p(o|\vec{\Theta})p(\vec{\Theta}|u') d\vec{\Theta}}.
\end{align}
}By dynamically and probabilistically estimating association hypothesis probabilities $\gamma$, the PSDA method enables more robust measurement updates by deriving a new mixture distribution, consisting of the posterior derived by naively fusing an observation $o$ and the prior by ignoring the observation $o$. In practice, since the number of mixands of $p_{PSDA}(\vec{\Theta}|o)$ is doubled (i.e. from $M$ to $2M$) at every measurement update, the computation cost can become problematic. To address this, a mixture reduction method such as Salmond's or Runnall's methods \cite{salmond1990,runnalls2007} is applied following each update. Also, when the softmax function is used as the observation likelihood, the posterior shown in (\ref{eq: posterior}) is analytically intractable. This can be handled by applying statistical approximation methods such as the Laplace approximation \cite{bishop2006} or variational Bayes importance sampling \cite{nisarTRO}. Algorithm \ref{alg: psda_cmab} summarizes the process for calculating the PSDA posterior for OA-CMABs. 
\begin{algorithm}[t]
{\footnotesize
\caption{PSDA measurement update \cite{wakayamaTRO} adapted for OA-CMABs} 
\label{alg: psda_cmab}
\begin{algorithmic}[1] 
\renewcommand{\algorithmicrequire}{\textbf{Input:}}
\renewcommand{\algorithmicensure}{\textbf{Output:}}
\REQUIRE Estimated weights, means, and covariances for $\vec{\Theta}_k$, a human semantic observation $o$, context vectors $\vec{x}_c$ and $\vec{x}_k$, the total number of possible observations $F$, and a prior faulty measurement probability (FP) rate $p(\zeta=0)$
\ENSURE Updated weights, means, and covariances for $\vec{\Theta}_k$
\FOR {each mixand $u$}
    \STATE $\vec{\mu}_{pos,u}$, $\Sigma_{pos,u}$, and $\mathcal{C}_{pos,u}$ via approximation algorithms \cite{bishop2006,nisarTRO} 
\ENDFOR
\STATE $\Lambda = \sum_{u=1}^{M} (w_{prior, u} \times \mathcal{C}_{pos, u} )$
\FOR {each mixand $u$}
    \STATE $w_{pos, u} = (w_{prior,u} \times \mathcal{C}_{pos,u}) / \Lambda$ 
\ENDFOR
\STATE $\gamma_0 = \frac{\frac{1}{F} \times FP}{\frac{1}{F} \times FP + (1-FP) \times \Lambda}$, $\gamma_1 = \frac{(1-FP) \times \Lambda}{\frac{1}{F} \times FP + (1-FP) \times \Lambda}$
\STATE $w_{prior} *= \gamma_0$, $w_{pos} *= \gamma_1$
\RETURN $w_{psda}, \vec{\mu}_{psda}, \Sigma_{psda}$ by stacking the prior and the posterior
\end{algorithmic}  
} 
\end{algorithm}\subsection{Expected Free Energy with Mixture Priors} 
As explained in Sec. \ref{sec: aif_efe}, it has become clearer in recent years that using active inference as an action/option selection strategy in MABs and CMABs can identify the best option with fewer iterations \cite{markovic2021,wakayamaICRA2023}. This is because by selecting an option minimizing expected free energy (EFE) agents can naturally balance exploitation and exploration while rigorously evaluating option uncertainties. Yet, in the previous approach for CMABs \cite{wakayamaICRA2023}, the prior distribution of hidden variables (here it is denoted as $\vec{\Theta}_k$) is {\it unimodal}; {\it multimodal} distributions such as those generally obtained via PSDA are not addressed. Thus, in the following, we present a more general derivation of EFE especially when a prior proposal distribution is a mixture pdf and the observation likelihood is the softmax function.

According to the active inference theory \cite{smith2022step}, the goal of a decision-making agent is to minimize the {\it surprise} of observations to maintain its homeostasis. The surprise in the case of OA-CMABs defined in Sec. \ref{sec: problem_statement} is expressed as 
{\allowdisplaybreaks
\abovedisplayskip = 2pt
\abovedisplayshortskip = 2pt
\belowdisplayskip = 2pt
\belowdisplayshortskip = 0pt
\begin{align} \label{eq: surprise}
    \mbox{Surprise} = -\log p(o) = -\log \int_{\vec{\Theta}} p(o, \vec{\Theta}) d\vec{\Theta}.
\end{align}
}However, calculating (\ref{eq: surprise}) directly via multiple integral tends to be analytically intractable, so its upper bound called {\it free energy} is tried to be minimized. Yet, in decision making, outcomes $o$ are unknown till an option is actually executed. Thus, the decision-making agent instead selects an option that minimizes EFE as shown in (\ref{eq: efe_general}). Hereafter, the option index $k$ and context vectors $\vec{x}_c$ and $\vec{x}_k$ are abbreviated for simplicity 
{\allowdisplaybreaks
\abovedisplayskip = 2pt
\abovedisplayshortskip = 2pt
\belowdisplayskip = 2pt
\belowdisplayshortskip = 0pt
\begin{align} \label{eq: efe_general}
    \mbox{EFE}(a)\!=\!\int_{\vec{\Theta}} q(\vec{\Theta}) \sum_o p(o|\vec{\Theta}) \log \frac{q(\vec{\Theta})}{p(\vec{\Theta}|o)p_{ev}(o)} d\vec{\Theta},
\end{align}
}where $q(\vec{\Theta})$ is a proposal prior and $p_{ev}(o)$ is an evolutionary prior, which reflects a (human's) prior preference for possible outcomes. Since the proposal prior $q(\vec{\Theta})$ is set as a multimodal distribution, (\ref{eq: efe_general}) is rewritten as follows.
{\allowdisplaybreaks
\abovedisplayskip = 2pt
\abovedisplayshortskip = 2pt
\belowdisplayskip = 2pt
\belowdisplayshortskip = 0pt
\begin{align} \label{eq: efe_general_mixture}
    \mbox{EFE}(a)\!=\!\int_{\vec{\Theta}} \sum_u q(\vec{\Theta}, u) \sum_o p(o|\vec{\Theta}) \log \frac{\sum_u q(\vec{\Theta}, u)}{p(\vec{\Theta}|o)p_{ev}(o)} d\vec{\Theta},
\end{align}
}where $q(\vec{\Theta}, u)$ is a joint proposal prior distribution. By applying the conditional dependency from the PGM and Bayes' rule, (\ref{eq: efe_general_mixture}) is further expanded as below. 
{\small
{\allowdisplaybreaks
\abovedisplayskip = 2pt
\abovedisplayshortskip = 2pt
\belowdisplayskip = 2pt
\belowdisplayshortskip = 0pt
\begin{align} 
    (\ref{eq: efe_general_mixture}) &= \sum_{o,u} q(u) \int_{\vec{\Theta}} q(\vec{\Theta}|u) p(o|\vec{\Theta}) \log \frac{q(o)}{p_{ev}(o) p(o|\vec{\Theta})} d\vec{\Theta},  \nonumber \\
    &= \sum_o \Big[ \sum_u q(u) \int_{\vec{\Theta}} q(\vec{\Theta}|u) p(o|\vec{\Theta}) \log \frac{q(o)}{p_{ev}(o)} d\vec{\Theta}, \nonumber \\
    &\ \ \ \ \ \ \ \ \ \ - \sum_{u} q(u) \int_{\vec{\Theta}} q(\vec{\Theta}|u) p(o|\vec{\Theta}) \log p(o|\vec{\Theta}) d\vec{\Theta} \Big] \label{eq: efe_mixture_details}.
\end{align}
}}where $q(o)$ is 
{\allowdisplaybreaks
\abovedisplayskip = 2pt
\abovedisplayshortskip = 2pt
\belowdisplayskip = 2pt
\belowdisplayshortskip = 0pt
\begin{align} \label{eq: qo}
    q(o) &= \int_{\vec{\Theta}} \sum_u q(\vec{\Theta}, u, o) d\vec{\Theta} = 
    \sum_u q(u) \int_{\vec{\Theta}} q(o|\vec{\Theta}) q(\vec{\Theta}|u) d\vec{\Theta}.
\end{align}
}The part of the first term of (\ref{eq: efe_mixture_details}) can be calculated by using the normalization constant $\mathcal{C}_u\!=\!\int_{\vec{\Theta}} q(\vec{\Theta}|u) p(o|\vec{\Theta}) d\vec{\Theta}$ when deriving the posterior for each mixand $u$, as when calculating association hypothesis probabilities in Sec. \ref{sec: update_sda}, 
{\allowdisplaybreaks
\abovedisplayskip = 2pt
\abovedisplayshortskip = 2pt
\belowdisplayskip = 2pt
\belowdisplayshortskip = 0pt
\begin{align} \label{eq: first_term}
    \mbox{(\nth{1})} = \log \frac{q(o)}{p_{ev}(o)} \sum_u q(u) \cdot  \mathcal{C}_u.
\end{align}
}However, the second term of (\ref{eq: efe_mixture_details}) cannot be calculated analytically because the integral of the log of the hybrid likelihood is analytically intractable. Nevertheless, if the prior $q(\vec{\Theta}|u)$ associated with each mixand $u$ is Gaussian, $\exp(\mathcal{L}_u\!+\!\mathcal{M}_u^T \vec{\Theta}\!-\! \frac{1}{2}\vec{\Theta}^T \mathcal{N}_u \vec{\Theta})$,  and the hybrid likelihood $p(o|\vec{\Theta})$ is softmax function, the posterior can be approximated as another Gaussian, $\exp (\mathcal{P}_u\!+\!\mathcal{Q}_u^T \vec{\Theta}\!-\!\frac{1}{2} \vec{\Theta}^T \mathcal{R}_u \vec{\Theta})$ \cite{nisarTRO}, so that the softmax function is approximately expressed as another exponential form, $\exp(\mathcal{G}\!+\!\mathcal{H}^T \vec{\Theta} \!-\!\frac{1}{2} \vec{\Theta}^T \mathcal{K} \vec{\Theta})$, where 
{\allowdisplaybreaks
\abovedisplayskip = 2pt
\abovedisplayshortskip = 2pt
\belowdisplayskip = 2pt
\belowdisplayshortskip = 0pt
\begin{align} \label{eq: backup_operation}
    \mathcal{G} &= \mathcal{P}_u + \log \Big( \int_{\vec{\Theta}} q(\vec{\Theta}|u) p(o|\vec{\Theta}) d\vec{\Theta} \Big) - \mathcal{L}_u, \\
    \mathcal{H} &= \mathcal{Q}_u - \mathcal{M}_u, \ \ \
    \mathcal{K} = \mathcal{R}_u - \mathcal{N}_u.
\end{align} 
}Note that this Gaussian approximation process for the softmax likelihood needs to be done {\it only once} from the fact of the conditional independence between $u$ and $o$ given $\vec{\Theta}$ (see the PGM in Fig. \ref{fig: pgm}). As a consequence, the part of the second term of (\ref{eq: efe_mixture_details}) is derived as follows, 
{\allowdisplaybreaks
\abovedisplayskip = 2pt
\abovedisplayshortskip = 2pt
\belowdisplayskip = 2pt
\belowdisplayshortskip = 0pt
\begin{align} \label{eq: second_term}
    (\mbox{\nth{2}}) = \sum_u q(u) \cdot \mathcal{C}_u \cdot \mathbb{E} [\mathcal{G} + \mathcal{H}^T \vec{\Theta} - \frac{1}{2} \vec{\Theta}^T \mathcal{K} \vec{\Theta}], 
\end{align}
}
and Algorithm \ref{alg: aif_cmab_mixture} outlines the process for calculating the EFE when mixture priors and hybrid likelihoods are used.
\begin{algorithm}[t] 
\caption{Generalized EFE calculation of \cite{wakayamaICRA2023} with mixture priors and hybrid likelihoods} \label{alg: aif_cmab_mixture}
{\footnotesize
\begin{algorithmic}[1] 
\renewcommand{\algorithmicrequire}{\textbf{Input:}}
\renewcommand{\algorithmicensure}{\textbf{Output:}}
\REQUIRE Estimated weights, means, and covariances of $\vec{\Theta}_k$, context vector $\vec{x}_c$ and $\vec{x}_k$, the total number of possible observations $F$, a prior faulty measurement probability (FP) rate $p(\zeta=0)$, and the evolutionary prior $p_{ev}(o)$  
\ENSURE EFE for selecting an option $k$
\FOR {each observation}
\STATE $\vec{\mu}_{pos, u}$, $\Sigma_{pos, u}$, and $\mathcal{C}_{pos, u}$ $,\forall u, u \in \{1, \cdots, M\}$ via approximation algorithms \cite{nisarTRO,bishop2006} 
\STATE $q(o)$ from (\ref{eq: qo}) and calculate the 1st term (\ref{eq: first_term})
\STATE $\mathcal{G}$, $\mathcal{H}$, and $\mathcal{K}$ from (\ref{eq: backup_operation}) and calculate the 2nd term (\ref{eq: second_term})
\STATE EFE$(a_k, o)$ = the 1st term - the 2nd term
\ENDFOR
\RETURN EFE$(a_k) = \sum_o \mbox{EFE} (a_k, o)$ 
\end{algorithmic} 
}
\end{algorithm}

\section{Simulation Study} \label{sec: simulation_study}
To validate the effectiveness of the proposed PSDA measurement update process and the generalized hybrid EFE calculation method for option selections for OA-CMABs, we performed a simulated asynchronous collaborative search site selection study modeled around the deep space exploration application shown in Fig. \ref{fig: scenario}. We first explain details for modeling the lander science exploration scenario as an OA-CMAB and describe the simulation experiment setup. Then, based on the results of the extensive Monte Carlo (MC) simulation, we at first establish which option selection approaches are effective for OA-CMABs by assuming that the FP rate is $0$. Thereafter, we examine how robust the PSDA method is to data association
uncertainties.

\subsection{Motivating Scenario}
\begin{figure}[t]
    \centering
    \centerline{\includegraphics[width=0.9\columnwidth]{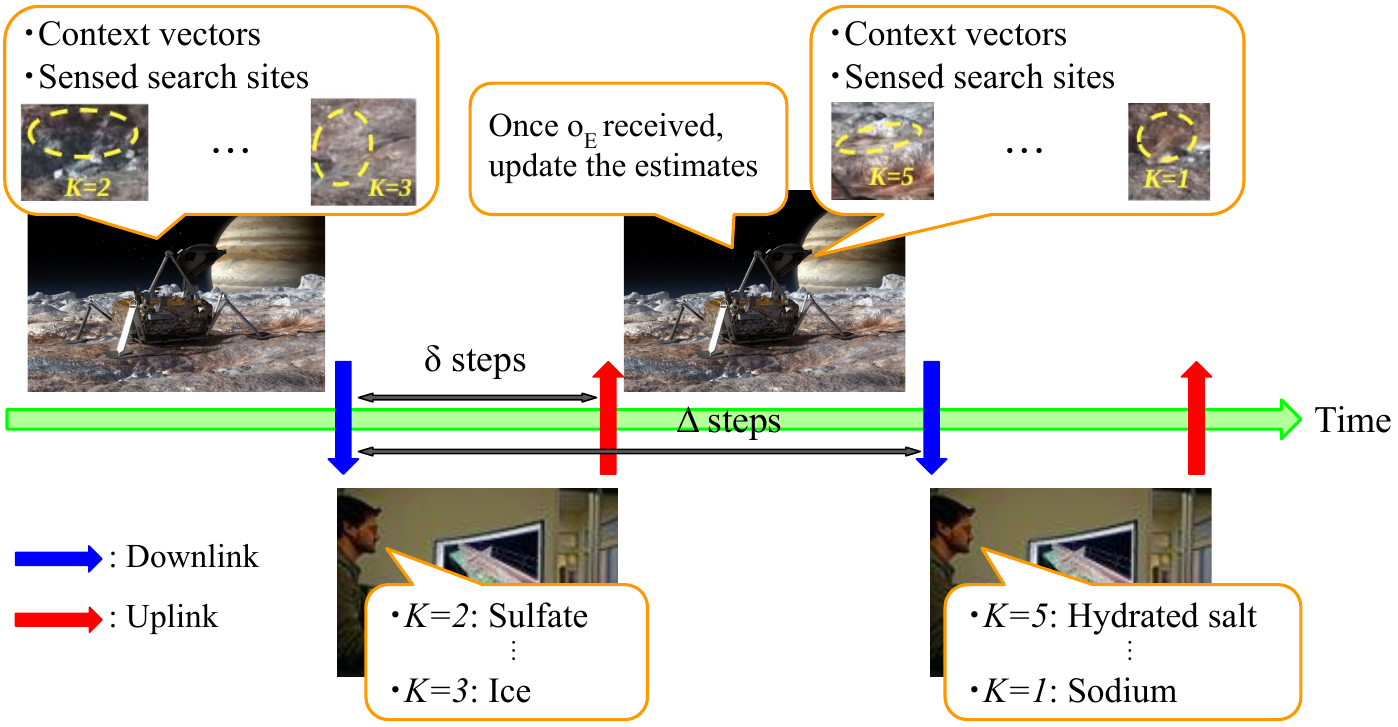}}
    \caption{{\small{Asynchronous lander and scientist communication.}}}
    \label{fig: comm}
    \vspace{-0.2in}
\end{figure}

Suppose the robotic lander is dispatched to icy moons like Europa or Enceladus \cite{hand2022science,aero2023}. One of the missions of the lander is to perform high-level science-related tasks such as deciding where to deploy a resource-intensive manipulator.
This can occur among the non-overlapping $K$ search sites that are determined based on panoramic images taken immediately after landing. To perform this science task, it is necessary to efficiently infer in advance, using lightweight sensors such as the LIBS \cite{jolivet2019review}, which search sites have the most features of interest $f_p$ to scientists out of the broadly classified $F$ scientific categories (e.g. hydrated salts and sulfates \cite{hand2022science}). While the lander can choose a site based solely on past observation data, it 
can utilize secondary context information to predict the likelihood of observing each outcome. For instance, contexts can include the sun direction, common to all search sites (i.e. $\vec{x}_c$), and albedo, which is unique to each search site (i.e. $\vec{x}_k$). Furthermore, as illustrated in Fig. \ref{fig: comm}, the lander periodically and asynchronously receives extra discrete semantic human observations about what scientists recognize in the downlinked data.  
Yet, the lander must also account for a non-zero probability that some observations are erroneous, e.g. due to the limited bandwidth and quality of data  transmitted 
back to Earth. 
We formulate this coupled inference and decision-making problem as an OA-CMAB.

\subsection{Simulation Setup} \label{sec: sim_setup}
The option selection problem for the OA-CMAB here corresponds to selecting a site to illuminate with LIBS, whereas measurement updating entails processing LIBS data returns and received human semantic observations. In this study, the following solution approaches are considered and compared in extensive MC simulation: (i) best possible option (site) selection, using an offline oracle (required to compute cumulative regrets); (ii) $\varepsilon$-greedy (where $\varepsilon = 0.25$ was found to work best after initial trials); (iii) upper confidence bound (UCB); (iv) multicategorical Thompson sampling (TS); and (v) active inference (AIF). The option selection methods for (iv) and (v) are paired with the Laplace approximations \cite{bishop2006} for the measurement updates. $100$ MC runs are performed, and the number of iterations $T$ in each MC run is set to $10^2$, which is much smaller compared to common MAB algorithm benchmarks \cite{markovic2021} and reflects a practical upper limit for robotic lander sensor deployment. Later in this section, we also analyze the asymptotic behaviors of the AIF and TS agents to discuss the pros/cons of using EFE. The true hidden linear parameters $\vec{\Theta}_{k}$ for each candidate search site $k$ were randomly generated from a uniform distribution of $0$ to $1$. The search-common and search-agnostic context vectors $\vec{x}_c$ and $\vec{x}_k$ was randomly generated assuming that each element takes a binary value with uniform probability. Note that the ways to generate $\vec{\Theta}_k$, $\vec{x}_c$ and $\vec{x}_k$ in this study may not align with actual phenomena since even partial environmental information is not available as of now. Yet, our approaches do not rely on any particular generative functions. Thus, once more suitable distributions are indicated from future deep space missions \cite{pappalardo2014}, these can be easily adapted. Finally, given the limited mission lifetime ($\approx$ 20 days \cite{adam2016europa}), it may be desirable to identify the best search site from one science perspective within several days (e.g. $3$ days) and move on to the next science objective. In particular, given the significant distance between icy moons and Earth -- resulting in a one-way communication time delay of approximately 45 minutes -- and the necessary margins for performing the other duties that the lander would have to do, the interval $\Delta$ at which it downlinks data is assumed to be $4$ steps (around $3$ hours), and the period $\delta$ between the downlink of data and the uplink of semantic 
information by the scientists is assumed to be $2$ steps (around $90$ minutes)\footnote{In our simulation experiments, for the most complex scenario ($K$=$15$, $C$=$3$, $F$=$12$), the maximum (i.e. using AIF) average computation time is 32.3 seconds, 
comporting with our assumptions for the communication intervals. It is worth noting that if there is prior knowledge indicating that only specific outcome observations are relevant, agents can limit the calculation of $EFE(a_k,o)$ to those outcomes, thus reducing computational costs. 
}. 
Our simulation experiments were conducted on a computer with Intel Core i7-8550 1.8 GHz 4-Core Processor.
\subsection{Results}
\subsubsection{Effectiveness of extra outcome observations from external information sources}
\begin{figure}[t]
    \centering
    \centerline{\includegraphics[width=0.9\columnwidth]{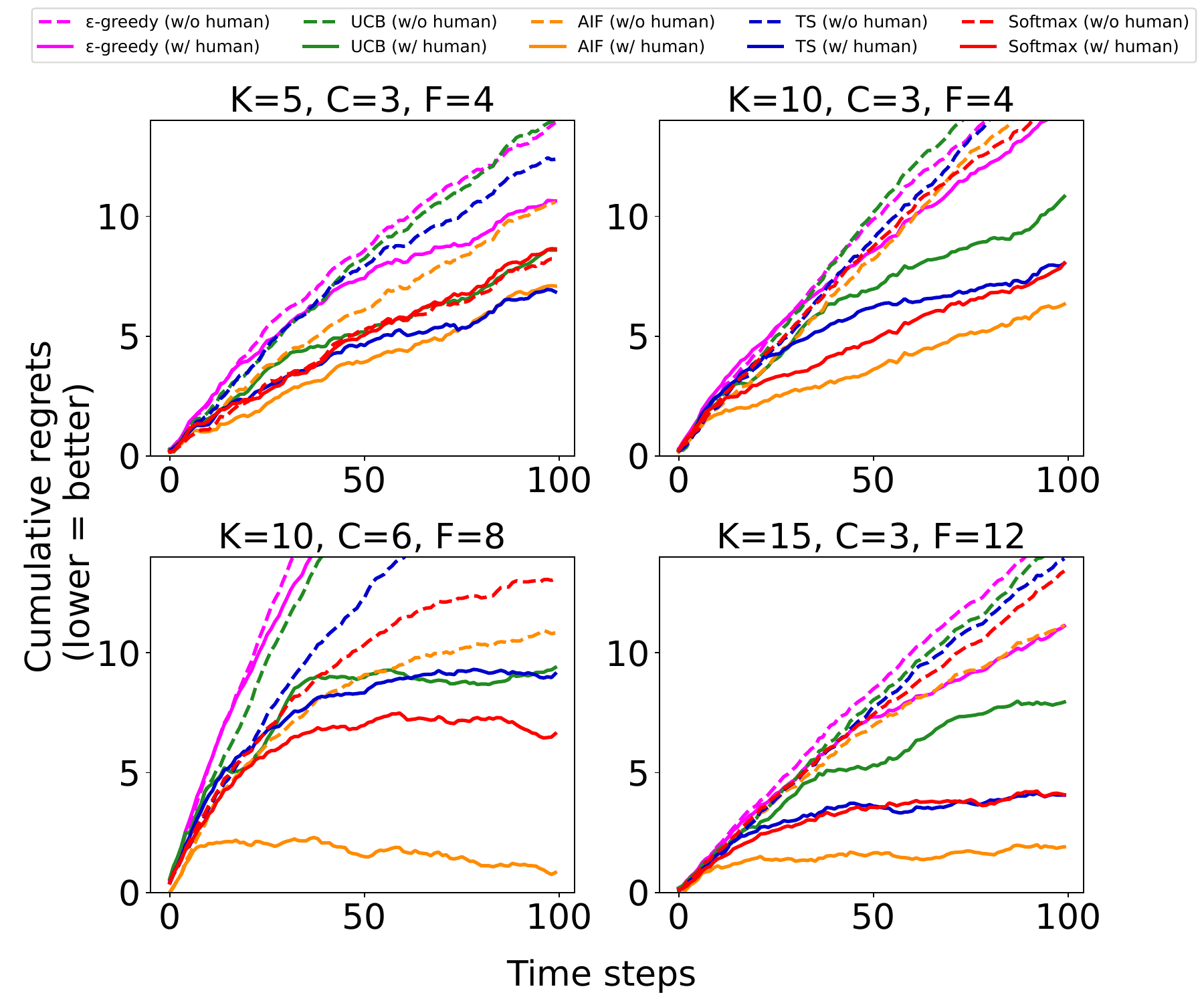}}
    \caption{{\small{Cumulative regrets when human semantic observations are always correct, i.e. $FP\!=\!0$.}}} 
    \label{fig: result_1}
    \vspace{-0.2in}
\end{figure}

\begin{figure}[t]
    \centering
    \centerline{\includegraphics[width=0.95\columnwidth]{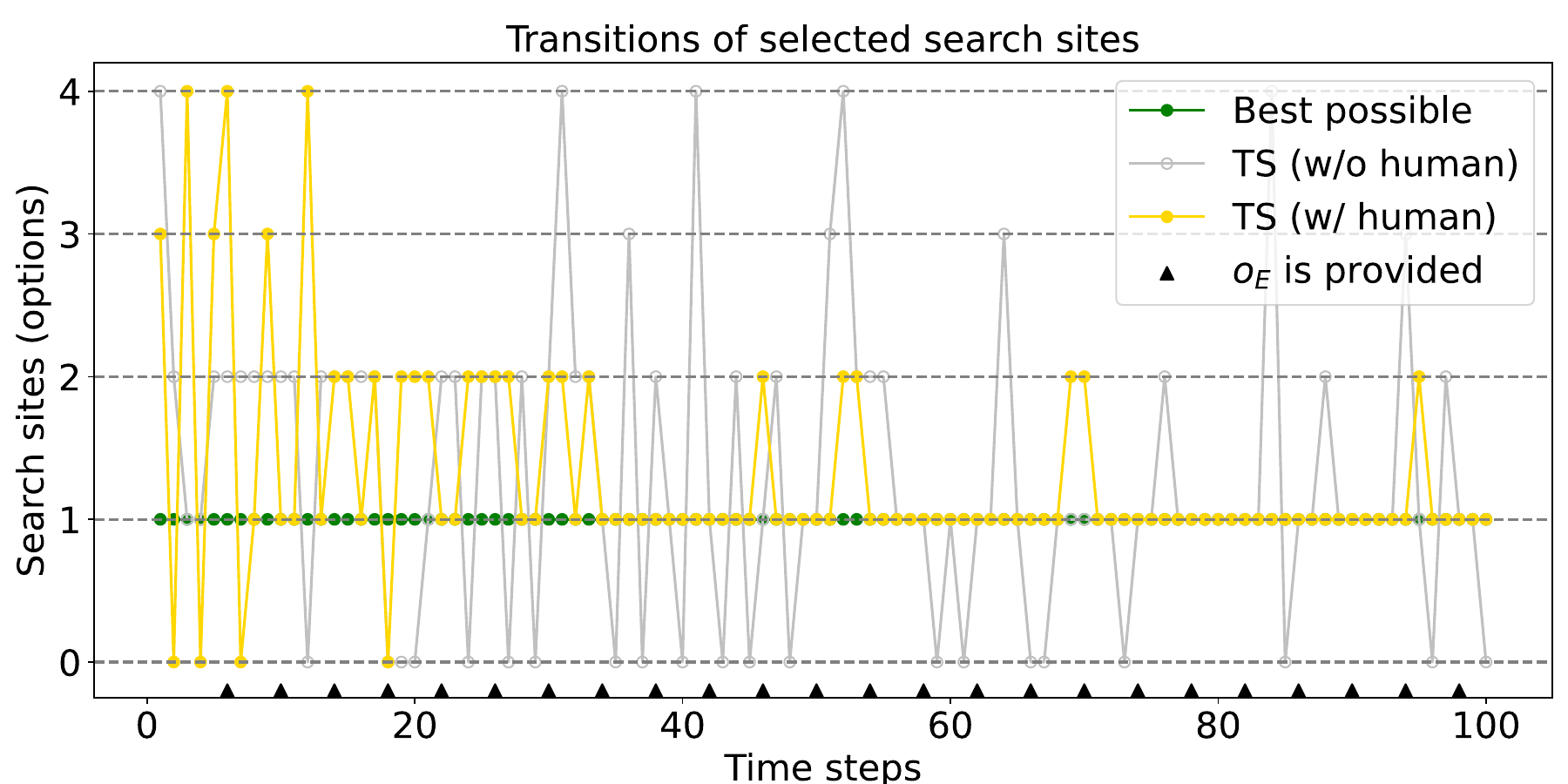}}
    \caption{{\small{Comparison of typical transition of selected search sites. Black triangles indicate fusion of external outcome observations $o_E$.}}}
    \label{fig: pulled_arm_transition}
    \vspace{-0.16in}
\end{figure}

In the first set of simulation experiments, we determine appropriate option selection approaches for OA-CMABs assuming no erroneous external observations. In the most difficult case (more options and features but less contexts), the total number of search site $K$ is $15$, the length of context vectors $C$ is $3$, and the number of science categories $F$ is $12$ (e.g. `Ice', `Hydrated salt', `Sulfate', and `Hydrated sulfuric acid'). Therefore, the lander is required to estimate $540$ hidden parameters in total. Assuming that scientists are most interested in hydrated sulfuric acid (i.e. it is $f_p$), the prior preference $p_{ev}(o)$ was set as $p_{ev}(o)\!=\!0.01$ if $o\!\neq\!f_p$. As a result, as can be seen in Fig. \ref{fig: result_1}, for all experiments, we observed that smaller cumulative regrets were achieved when (asynchronous) human semantic information was used to estimate the latent parameter vectors (solid lines) than when it was not used (dashed lines). This is because, as illustrated in Fig. \ref{fig: pulled_arm_transition}, compared to the approach without extra external observations $o_E$ (gray), the robot using the approach with additional observations $o_E$ (yellow) selects the best search site ($k\!=\!1$ for this instance) much more often. Since the results were notably good when AIF and TS were used with human semantic observations (orange and blue plots in Fig. \ref{fig: result_1}), only these two approaches are used in the rest of the experiments. Also, hereafter we focus on the case of $K\!=\!5$, $C\!=\!3$, and $F\!=\!4$, since compared to the other cases, the alternative methods considered here become most competitive with AIF.

\subsubsection{Performance of the PSDA algorithm}
\begin{figure*}[t]
    \centering
    \includegraphics[width=0.85\linewidth]{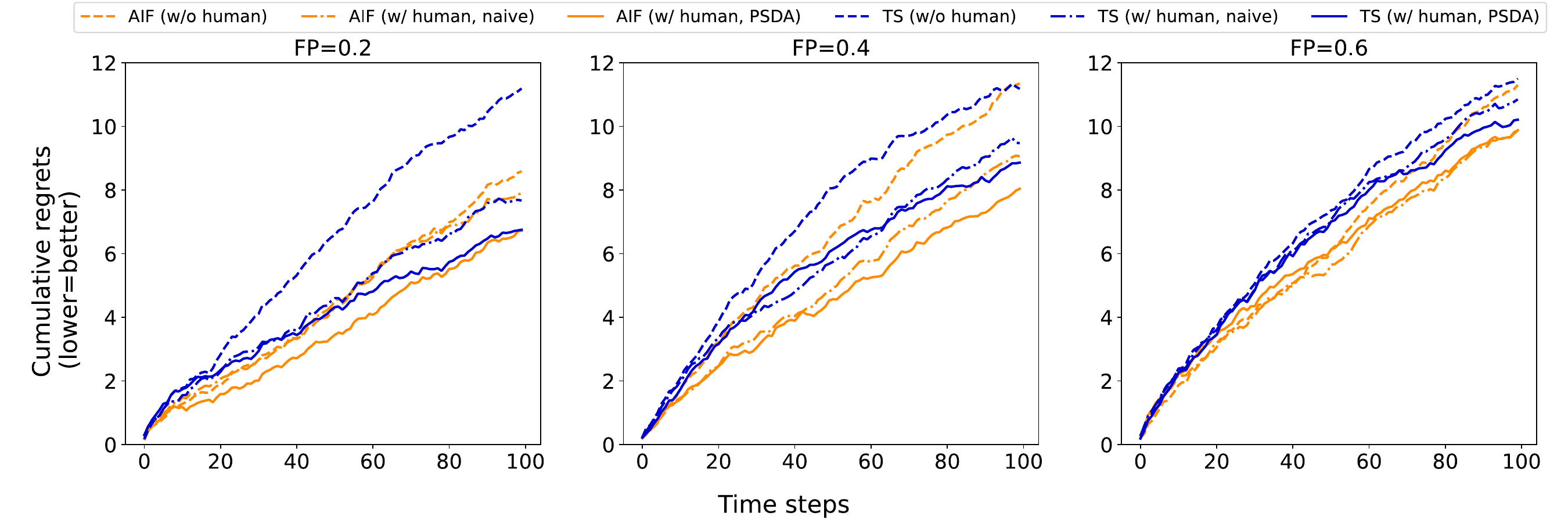}
    \caption{{\small{Cumulative regrets with different data fusion modalities and the FP rates ($0.2$, $0.4$, and $0.6$). Even if the FP rate is high enough, when the PSDA measurement update is used, the cumulative regrets are lower compared to the other modalities.}}}
    \label{fig: varying_FP}
\vspace{-0.15in}
\end{figure*}

\begin{figure}[t]
    \centering
    \includegraphics[width=0.88\linewidth]{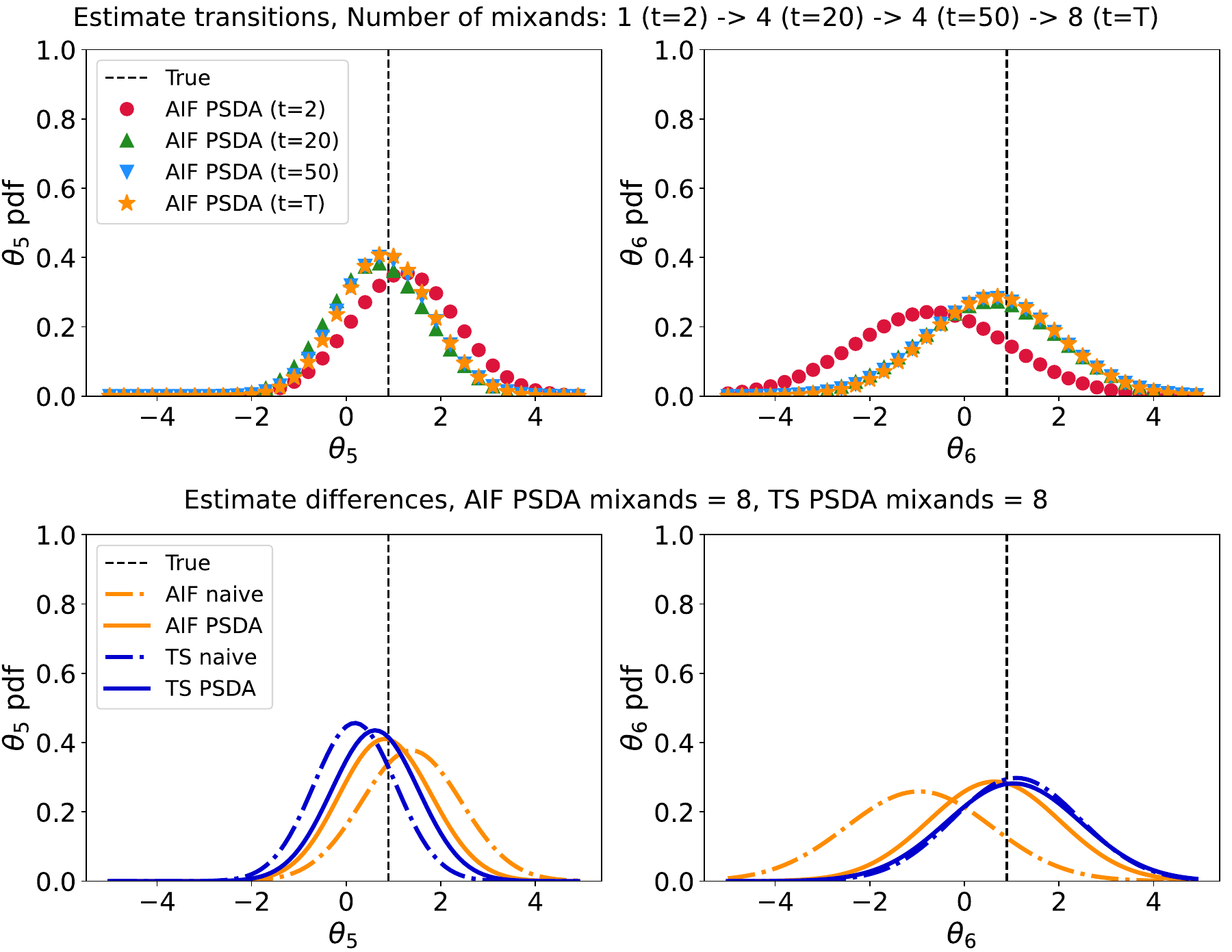}
    \caption{{\small{Typical estimate transitions of the AIF PSDA marginal posteriors for hidden parameters $\theta_5$ and $\theta_6$ (true values at dashed vertical lines) for $K\!=\!2$, $FP\!=\!0.4$, 
    and $t \in \{2, 20, 50, 100\}$ for a single Monte Carlo run. 
    (top row): typical final marginal posteriors at same conditions for different data fusion methods (bottom row).}}} 
    \label{fig: estimates_transitions}
    \vspace{-0.2in}
\end{figure}

Next, to assess the benefit of the proposed PSDA measurement update algorithm, we conducted simulation experiments with the prior probability of human semantic observation being incorrect, i.e. $FP\!=\!p(\zeta\!=\!0)$, set at $0.2$, $0.4$, and $0.6$\footnote{Previous work \cite{wakayamaTRO} showed that autonomous agents do need to know the precise FP rate, 
as long as the assumed FP rate is higher than the true rate.}. Here we compared the following data fusion modalities 1) no semantic observations (w/o human), 2) naively fuse semantic observations (w/ human, naive), and 3) fuse semantic observations with PSDA (w/ human, PSDA), for AIF and TS. The threshold value for activating the Runnall's mixture reduction method was set as $10$. As can be seen from Fig. \ref{fig: estimates_transitions} (top row), even when the human semantic observations were occasionally erroneous, by dynamically and probabilistically estimating association hypothesis probabilities $\gamma$, the PSDA estimates gradually approached to the true hidden parameters (dashed vertical lines) and the final estimates were better compared to the ones when incorrect human semantic observations were naively fused as in Fig. \ref{fig: estimates_transitions} (bottom row). As a consequence, employing the PSDA measurement update algorithm resulted in smaller cumulative regrets (Fig. \ref{fig: varying_FP}). Note that in Fig. \ref{fig: estimates_transitions}, the marginal posteriors look nearly unimodal/Gaussian, as various mixands cluster near the same locations for the observations provided, though some pdfs are actually skewed/asymmetric. Such clustering will not always occur, e.g. when humans provide negative observations (not used here) to scatter posterior mixand locations \cite{wakayamaTRO}.   
\subsubsection{Asymptotic behaviors of AIF and TS agents} 
\begin{figure}[htbp]
    \centering
    \centerline{\includegraphics[width=0.67\columnwidth]{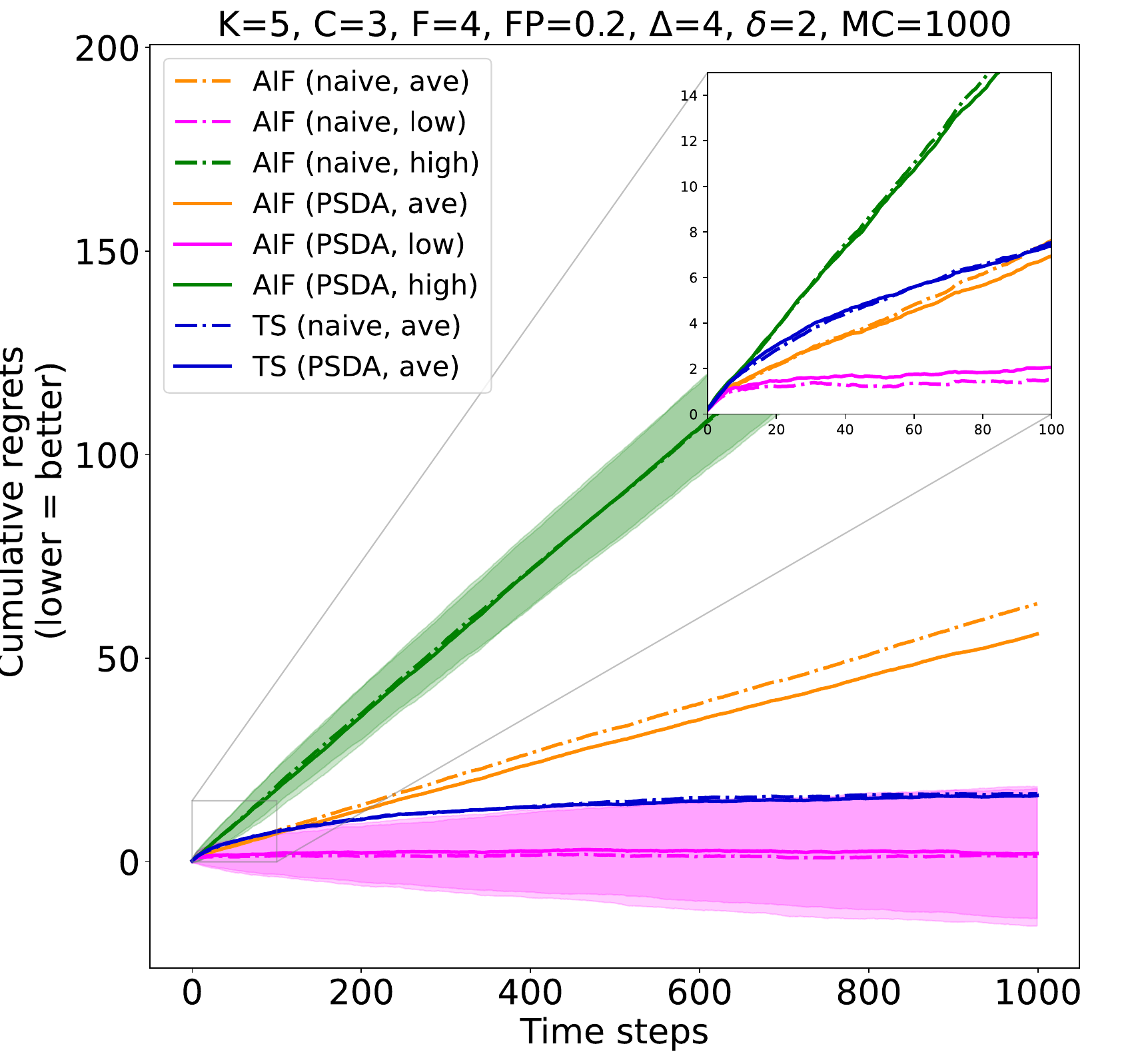}}
    \caption{{\small{Asymptotic cumulative regrets for AIF and TS. In the early stages, AIF outperform TS. Yet, due to the bimodality of AIF (magenta and green lines), the average asymptotic behaviors (orange lines) are worse than TS (blue lines).}}}
    \label{fig: asymptotic}
    \vspace{-0.2in}
\end{figure}

\begin{figure}[htbp]
    \centering
    \centerline{\includegraphics[width=0.67\columnwidth]{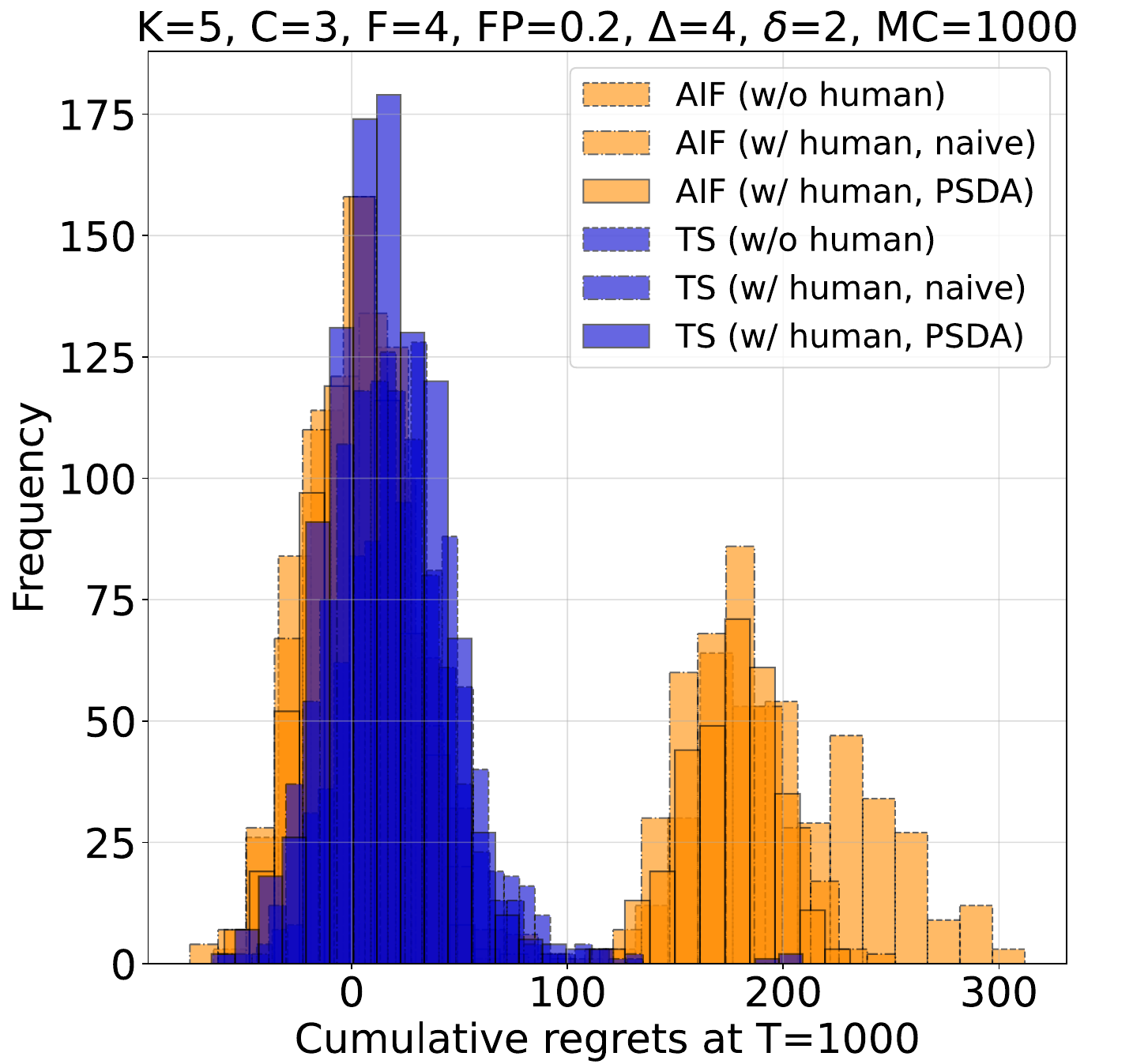}}
    \caption{{\small{Cumulative regrets at $T\!=\!10^3$. As with standard MABs \cite{markovic2021}, 
    the AIF agents exhibit their tendencies to be stuck at local minima and produce the bimodal cumulative regrets.}}}
    \label{fig: hist}
    \vspace{-0.2in}
\end{figure}

The results so far show that (under the same simulation conditions) AIF outperforms TS. However, in stationary MABs, it is experimentally known that the asymptotic behaviors of AIF can be worse than TS, due to the biased generative model by incorporating evolutionary prior $p_{ev}(o)$ \cite{markovic2021}. Thus, we conducted another simulation experiment with $MC$ and $T$ set to $10^3$ each to see if the same trend could be confirmed for contextual bandits. As shown in Fig. \ref{fig: asymptotic}, initially, the AIF agents (orange lines) perform better than those of the TS agents (blue lines). However, as Fig. \ref{fig: hist} shows, when the AIF agents find one of multiple good sites where a desired outcome can be easily observed (not necessarily the best one), the EFE value of this site can be smaller than the others due to the very small exploitation term, leading 
to the cluster on the right-hand side of this figure. Thus, even if one subset of AIF agents quickly determines the best search site and generates very small cumulative regrets (magenta lines in Fig. \ref{fig: asymptotic}), the suboptimal behavior of the remaining AIF agents (green lines in Fig. \ref{fig: asymptotic}) causes average cumulative regrets (orange lines)  to gradually become larger than those of the TS agents. Therefore, the AIF-based option selection method is generally more suitable for switching bandit problems where the underlying (reward) models associated with options change dynamically \cite{jun2004survey, gupta2011thompson}.

\section{Conclusion and Future Works} \label{sec: conclusions}
We introduced observation-augmented contextual multi-armed bandits to utilize semantic observations from external sources that can accelerate parameter inference for robotic decision making. To address 
possible errors
in such observations, we developed a robust Bayesian inference process to dynamically evaluate data validity. We also derived a generalized expected free energy approximation for active inference option selection with mixture-based parameter priors and 
observation likelihoods inherent to semantic data validation problems. Simulation studies showed our methods achieve smaller cumulative regrets vs. other conventional
bandit 
algorithms, even with erroneous external observations. 
One next step is to validate our methods on more realistic deep space simulation environments with humans on the loop. In the case of icy moon site selection, the required data may be obtained from 
a high-fidelity physics simulator such as OceanWATERS \cite{oceanwaters}. Additionally, in our 
experiments, 
$p_{ev}(o)$ was fixed across all iterations, which may not be true in actual operations, e.g. as scientists may change their preferences to study new phenomena. Thus, it is also worthwhile to 
consider
dynamic $p_{ev}(o)$ for active inference.

\bibliographystyle{IEEEtran}
\bibliography{root}

\end{document}